# AI Fairness for People with Disabilities: Point of View

Shari Trewin, IBM Accessibility Research, trewin@us.ibm.com

## Abstract
We consider how fair treatment in society for people with disabilities might be impacted by the rise in the use of artificial intelligence, and especially machine learning methods. We argue that fairness for people with disabilities is different to fairness for other protected attributes such as age, gender or race. One major difference is the extreme diversity of ways disabilities manifest, and people adapt. Secondly, disability information is highly sensitive and not always shared, precisely because of the potential for discrimination. Given these differences, we explore definitions of fairness and how well they work in the disability space. Finally, we suggest ways of approaching fairness for people with disabilities in AI applications.

## Introduction

Artificial intelligence (AI) is everywhere. One very successful form of AI, machine learning (ML) models, are already helping doctors to spot melanoma, recruiters to find promising candidates, and banks to decide who to extend a loan to. They are used in product recommendations, targeted advertising, essay grading, employee promotion and retention, image labelling, video surveillance, self-driving cars and a host of other applications. Some of these applications have high stakes consequences for the individuals involved.

AI is also transforming the way we interact with machines, as they learn to translate our words to text, interpret our gestures, and recognize us and our emotions. Speech recognition has reached near-parity with human performance on some data sets (Xiong et al., 2017), and AI methods are able to identify people and objects in photos, video or sensor data. Autonomous vehicles are already out there.

Machine learning models are trained by examining large numbers of examples. They learn patterns in the training data, using these to classify new examples. Much has been written about the potential for these models to encode, perpetuate and even amplify discrimination against marginalized groups in society. Researchers have found gender discrimination, racial discrimination, and age discrimination in ML models. Disability discrimination has not been explored in the literature to date but there is clear potential for bias against people with disabilities in AI systems.

Like age, gender and race, disability status is a protected characteristic. In the United States, the Americans with Disabilities Act (ADA) of 1990 prohibits discrimination on the basis of disability in employment, access to government programs and services, public transportation, public accommodations and telecommunications. In Europe, disability is recognized as a grounds for discrimination, and the Framework Directive protects people with disabilities from discrimination in employment and occupation (Council of the European Union, 2000). The United Nations Convention on the Rights of People with Disabilities (UN General Assembly, 2007) has been ratified by 177 countries.

This paper describes the unique characteristics of disability as a protected characteristic, discusses the suitability of several common definitions of fairness, and offers recommendations for addressing fairness for people with disabilities in AI solutions.

## Fairness in AI Systems

As AI models become pervasive, it is essential that they uphold society's moral and legal obligations to treat citizens fairly, especially with respect to protected groups that have historically experienced discrimination. AI-based applications should not embody bias against any protected group. However, experience has shown that such bias can and does exist. One well-known example is the COMPAS model used to predict recidivism - the likelihood that a prisoner, if released, will commit a crime. By comparing against actual crimes committed, COMPAS was shown to have a racial bias, in which it was more likely to wrongly deny release to a black person, in comparison to a white person (Angwin, Larson, Mattu, & Kirchner, 2016).

Systematic bias can arise if data used to train a model contains human decisions that are biased, and the bias is passed on to the learned model. For example, if college recruiters systematically overlook applications from students with disabilities, or a health insurer routinely denies coverage to people with disabilities, a model trained on that data will replicate the same behavior.

Another source of bias can be lack of representation in data sets. For example, prominent face analytics systems were found to have much higher error rates for black women than white men (Buolamwini & Gebru, 2018). When this problem was identified, IBM moved quickly to correct the bias in its system by balancing the training data, reducing the disparity by an order of magnitude (Puri, 2018). Using representative data is an essential step to addressing potential bias against people with disabilities. The US Census Bureau estimates that 1 in 5 people have some form of disability in the U.S. (Brault, 2012). If data sets include this population, the resulting models are more likely to be effective for them.

Finally, bias can arise when the true quantity of interest is not directly measured, and other data is used as a proxy (Eubanks, 2018; O'Neil, 2017). Eubanks (Eubanks, 2018) describes the Alleghany County Screening Tool, a system for predicting risk of child abuse that makes heavy use of attributes relating to usage of public assistance programs as predictive features (Eubanks, 2018, p156). These measures are associated with poverty, leading to a bias against poorer families in the model. A similar situation can easily arise for disability. For example if the time taken to complete an online test is interpreted as reflecting the test taker's level of skill, this will disadvantage people using assistive technologies to access the test, especially if the test has not been made fully accessible.

## Fairness for People with Disabilities

Is fairness for people with disabilities any different to fairness for other protected groups, based on gender, race, age, or other attributes? We argue that in many cases it is. There are some crucial differences between disability groups and other protected groups.

### Diversity and Outliers

The greatest difference is the diverse, nuanced, and dynamic nature of disability itself. Disability is not a simple variable with a small number of discrete values. It has many dimensions and people can experience multiple disabilities. We define disability in line with the United Nations Convention on the Rights of People with Disabilities (CRPD) as a mismatch between the available infrastructure and the needs of an individual. As described in the CPRD, disability is not an attribute of a person, but "an evolving concept and … results from the interaction between persons with impairments and attitudinal and environmental barriers that hinders their full and effective participation in society on an equal basis with others." (UN General Assembly, 2007) Impairments and health conditions that can lead to disabilities are not only diverse, but vary in intensity and impact, and often change over time. This diversity also applies within groups that might at first seem homogeneous. A popular saying in the autism community is: "When you've met one person with autism, you've met one person with autism." Rather than forming a cohesive group, the disabled community includes many outliers.

This poses a challenge for machine learning, which works by finding patterns and forming groups. Outlier data is often treated as 'noise' and disregarded. Even with the data included, there may not be enough individuals with a given type and severity of disability in a data set for the machine to identify a pattern. Including outlier data makes the learning task more difficult and leads to more complex models that may overfit the data. Simpler models are preferred, and their predictions for these 'outlier' individuals may be poor quality, or unfairly negative. Compared to gender, race or age, it is not easy to address biased outcomes by gathering a balanced set of training data, because there are so many forms and degrees of disability.

### Privacy

Many people have privacy concerns with sharing their disability information. People with disabilities do experience discrimination and exclusion in everyday life. In one recent field study, disclosing a disability (spinal cord injury or Asperger's Syndrome) in a job application cover letter resulted in 26% fewer positive responses from employers, even though the disability was not likely to affect productivity for the position (Ameri et al., 2018). Another study found that the pay gap between people with and without disability cannot be explained by productivity differences, and likely represents discrimination (Kruse, Schur, Rogers, & Ameri, 2017).

Individuals rightly wish to have control over what is known about them and how that information is used. As a case in point, recently it came to light that the ACT standardized testing organization in the US was not only passing disability information about college applicants on to colleges, but even selling it to third parties (Jaschik, 2018). Students with disabilities are bringing a lawsuit against ACT for illegally revealing their disabilities to colleges they are applying to. For colleges to use this information in making admission decisions would also be illegal under the Americans with Disabilities Act (ADA).

In Europe, the new GDPR regulations give individuals the right to know what data about them is being kept, and how it is used, and to request their data to be deleted. As organizations move to limit the information they store and the ways it can be used, this means that AI systems may often not have explicit information about disability that can be used to apply established fairness tests and corrections.

## Approaches to Fairness

### Algorithmic Fairness

When AI approaches are used to support communication based on speech, writing, or gestures, the primary fairness concern is algorithmic fairness. The models should work equally well for members of different groups. For example, a speech recognition system should work as well for women's speech as it does for that of men. Depending on the application, AI methods may be inaccurate, or simply not work at all for some individuals because their appearance, speech or behavior are outside the AI's training data. There is anecdotal evidence from deaf speakers, for example, that today's speech recognition systems have very high error rates for understanding deaf speech. People who have a speech impairment are often unable to use such systems.

This aspect of fairness can be improved by gathering training data from a broad set of groups, and ensuring the process of 'cleaning' the data retains enough diversity. However, it is possible that using more diverse training data can degrade the overall performance of some models. In such cases it may be necessary to build specialized models for known groups, such as recognition of deaf speech. Researchers should explore methods to correctly handle data generated by outlier individuals and groups who speak, write, look, or behave differently from the average person.

### Allocative Fairness

For algorithms that allocate people to favored or less favored groups, many researchers have explored different ways of measuring and ensuring fairness in how the favorable outcomes are assigned (Dwork, Hardt, Pitassi, Reingold, & Zemel, 2012; Verma & Rubin, 2018).

What definition of fairness should be preferred in the disability space? Here, we address this question by examining some of the commonly used approaches, and their strengths and weaknesses. As a running example, imagine a job posting with 1000 applicants, 50 of whom are people with disabilities. All applicants are asked to take an online test of their knowledge of the job domain, and their results, resumes and application letters are used by an AI-based decision system to suggest a shortlist of 20 applicants to interview.

#### Fairness through unawareness

The simplest approach to fairness is 'fairness through unawareness', where no information about protected attributes (e.g. gender, age, disability) is gathered and used in the decision-making. This can be effective in some situations and aligns with some existing practice. For example, many job seekers choose not to reveal disabilities in initial job applications, and the Americans with Disabilities Act (ADA) specifically prohibits prospective employers from asking about applicants' disabilities.

However, this is not a guarantee of fairness. It can fail when an applicant's disability impacts other information used in the decision. For example, if five of our job applicants use assistive technologies such as a screen reader or magnifier, and the online test itself is not fully accessible, then long response times could lead to systematic exclusion of these five applicants using assistive technologies, even though their disability is not known.

#### Fairness through awareness

Without disability information in the data, it is difficult to assess whether systematic discrimination is occurring. As a result, AI researchers have advocated for 'fairness through awareness' (Dwork et al., 2012), where membership in a protected group is explicitly known, and fairness can be formally defined, tested and enforced algorithmically.

When disability information is available, data bias can be tackled with numerical methods, and the output of models can be adjusted to mitigate bias. In some applications, outcome information can be used to assess accuracy of predictions made by models for different groups (Gajane & Pechenizkiy, 2017; Verma & Rubin, 2018). IBM's [AI Fairness 360 Toolkit](#) (Bellamy et al., 2018) contains an extensive set of methods applicable to situations with and without outcomes data, offering a range of different fairness measures that can be used to identify bias, and assess the effectiveness of bias mitigation steps.

### Measuring Fairness - Group Fairness

One way to measure fairness when information is available is group fairness, or *statistical parity*. Under this policy, the proportion of selected individuals in a protected group (people with disabilities) should be roughly the same as the proportion of selected individuals in the non-protected group. In other words, the 20 applicants selected for interview should include one applicant with a disability. This is an affirmative action policy, but does not guarantee that the most qualified candidates will be selected.

Group fairness can be difficult to apply to disability:
1. It relies on having explicit disability information. As described above, this information can be highly sensitive, and because of existing legal protections, it may not be available (as in our job application example).
2. Testing at an aggregated level may conceal discrimination against specific sub-groups (for example those using assistive technologies for the aptitude test). Testing all possible disability sub-groups, on the other hand, is impractical, and there will rarely be enough data points or detailed disability information to do this,
3. It cannot accommodate genuine differences between groups.

### Measuring Fairness - Individual Fairness

An alternative approach is 'individual fairness', which specifies that similar individuals should have similar outcomes. This approach uses a metric to define how similar two individuals are with respect to the task at hand (Dwork et al., 2012). If an appropriate (and fair!) way to measure similarity can be defined, fairness can be tested using this metric. However, this approach poses the challenge of defining a fair similarity metric, and the question arises of whether and how disability information is included in such a metric. A fair similarity metric for the job application example might use measures of relevant experience and test scores, but not use protected information like disability status. Thus, applicants with similar test scores and qualifications should be equally likely to be selected, whether or not they have a disability. This has the advantage that fair treatment of any individual can be tested without knowing whether or not they have a disability, by comparing outcomes for 'similar' applicants. Test suites representing a broad set of people with disabilities could be used to probe for fairness in the resulting models.

The challenge with this approach is that every application needs its own well-defined and fair similarity metric. Bias can easily be introduced to such a metric by using attributes that are impacted by disability, such as test taking time.

## Ensuring Fairness

In technology access, we rely on standards to mediate between the variety of different needs, and the steps a developer must take. Developers of machine learning solutions may need similar assistance to help them understand and assess the impact of their applications for people with disabilities.

To improve fairness for people with disabilities in AI-based solutions, we propose that developers of AI-driven solutions that use data about people should:
- Consider the implications of their solution for people with disabilities, asking questions including: Who would present unusual data patterns that might not be handled well by a machine learning approach trained on typical data? Is there a disability group that stands to be severely affected? Including members of such groups in application design and development, and gathering training data specifically from these groups wherever possible can help to address concerns.
- Use existing techniques to test for bias and mitigate bias throughout the machine learning pipeline (in data, in models, and in outcomes), when disability information is available. The IBM AI Fairness 360 toolkit, for example, includes a range of fairness measures and techniques for pre-processing data to reduce bias, training fair models, and processing model output to improve fairness.
- Consider whether other variables in the input data might also be impacted by disability. A common example would be the time taken to perform a task. For applications in this class, an 'individual fairness' approach may be an appropriate way to define and measure fairness.
- Machine learning approaches work well for typical inputs, and do not handle outliers well. Although some aspects of fairness for people with disabilities could be improved by increasing representation in training data, there will always be outlier individuals. Fairness for people with disabilities, and robustness of machine learning solutions in real-world applications, could be improved by developing approaches that recognize and handle outliers.
- Finally, document the service so that others can assess fairness for people with disabilities. This includes: whether people with disabilities have been included in data collection, design and testing; the results of fairness testing; any steps taken to mitigate bias against people with disabilities; and known limitations of the service for specific disability groups.

## Conclusion

Although much has been written about gender, racial and age bias in AI systems, the potential for reducing or entrenching disability discrimination has received little attention in the research literature to date. Treviranus (Treviranus, 2017) highlights potential dangers of relying on pure ML approaches for applications that must work for everyone, not just the average person. The typical training process for ML models optimizes performance for typical cases, at the expense of edge cases, yet people with disabilities are very diverse, and may often appear as edge cases in the data.

For systems that will make or influence decisions affecting human lives, it is critical that a broad range of user stakeholders are involved in development, including people with disabilities who can help developers to think through the possible implications of the technology, and to test the technology's performance on edge cases and under-represented populations.

## Acknowledgments

Many thanks to Erich Manser, Phill Jenkins, Peter Fay, Lorene Southworth, Marc Johlic, Belal Oboabdo and Simeon McAleer of the IBM Accessibility Research Leadership Team for insightful discussions and research contributions that are reflected in this paper.